# *MRecover*: A Conditional Generative Model for Recovering Motion-Corrupted MR images Using AI Generated Contrast

Jinghang Li[#1], Tales Santini[#1], Courtney Clark[2], Bruno de Almeida[1], Cong Chu[1], Salem Alkhateeb[1], Andrea Sajewski[1], Jacob Berardinelli[1], Hecheng Jin[1], Tobias Campos[1], Jeremy J. Berardo[1], Joseph Mettenburg[3], Ariel Gildengers[4], Howard Aizenstein[4], Minjie Wu[4], Tamer S. Ibrahim[1,3,4*]

Affiliations:
[1] Department of Bioengineering, University of Pittsburgh, Pittsburgh, Pennsylvania, USA
[2] School of Medicine, University of Pittsburgh, Pittsburgh, Pennsylvania, USA
[3] Department of Radiology, University of Pittsburgh, Pittsburgh, Pennsylvania, USA
[4] Department of Psychiatry, University of Pittsburgh, Pittsburgh, Pennsylvania, USA

[#] Contributed equally and share the first authorship

[*] Correspondence:
Tamer S. Ibrahim, PhD
Professor of Bioengineering, Psychiatry, and Radiology
Swanson School of Engineering, and School of Medicine
University of Pittsburgh
3501 Fifth Avenue, Pittsburgh, PA 15213
tibrahim@pitt.edu

## Abstract
Hippocampal subfield segmentation requires high-resolution T2w turbo spin echo (TSE) MRI, yet this sequence is susceptible to motion artifacts, leading to substantial data loss. We developed a conditional generative model (*MRecover*) that synthesizes routinely acquired T1w images to create TSE images with autoregressive slice conditioning for volumetric consistency. Trained on 7T MRI data (n=577), the model achieved high in-domain fidelity (n=148, SSIM=0.84, FSIM=0.94) and generalized well to out-of-domain 3T data: subfield volumes from synthesized and the as-acquired images closely matched: (n=416, r=0.87–0.97) and yielded 31.8% more analyzable subjects in the motion-affected ADNI3 dataset after quality control (593 vs 450). The synthesized images also achieved larger effect sizes due to increasing the sample size for diagnostic group differences in hippocampal subfield atrophy (whole hippocampus $\varepsilon^2$=0.121–0.100 vs. 0.086–0.062, left-right hemispheres). Project page: https://jinghangli98.github.io/MRecover/

Keywords: Hippocampus subfield analysis; MRI contrast synthesis; 7T/3T.

## Introduction

The hippocampus is a structurally complex subcortical region highly involved in episodic memory, spatial processing, and learning[1-4]. Its subfields, including cornu ammonis (CA1, CA2/CA3), dentate gyrus (DG), subiculum (Sub), and entorhinal cortex (ErC), show distinct patterns of vulnerability across neurodegenerative and neurological disorders[5,6]. Accurate segmentation of these subfields is therefore essential for their use as imaging biomarkers. Because these subregions are small and tightly folded, resolving them with in-vivo imaging requires both high spatial resolution and tissue contrast that clearly depicts internal hippocampal anatomy.

High-resolution T2-weighted turbo spin echo (T2w-TSE) imaging has proven especially valuable for hippocampal subfield delineation and is now a standard modality for hippocampus subfield analysis[7,8]. Many open-source neuroimaging tools for hippocampal subfield segmentation, including FreeSurfer[9], Automatic Segmentation of Hippocampal Subfields (ASHS)[10], and HippUnfold[11], are specifically designed to use T2w-TSE acquisitions for more accurate segmentation. Subfield volume estimates derived from T2w-TSE can also provide greater Alzheimer's Disease (AD) diagnostic sensitivity than whole-hippocampal measures alone[12,13]. Additionally, subfield-specific atrophy patterns have been validated as biomarkers for AD progression, mesial temporal sclerosis, and age-related memory decline[14-16].

Despite its clinical value, T2w-TSE is fundamentally limited by its sensitivity to patient motion. Its high in-plane resolution and multi-shot Cartesian k-space acquisition make it particularly vulnerable to motion artifacts. Even submillimeter displacements can introduce ghosting and blurring artifacts that can obscure hippocampal subfield boundaries[17,18]. This limitation is most prominent in populations whose subfield analysis matters most: elderly adults, Alzheimer's disease patients, and individuals with cognitive impairment. These groups are disproportionately prone to in-scanner motion and face motion-related quality control exclusions in clinical neuroimaging studies[19-21]. Critically, this data loss is not random. Subjects excluded for motion are often those most likely to exhibit disease-related anatomical abnormalities, introducing selection bias and potentially attenuating group-level differences[22]. Existing motion mitigation strategies, including prospective motion correction[23,24], and PROPELLER-based acquisition[25,26], can partially reduce artifact burden during acquisition, but they are difficult to implement and cannot recover data that are excluded after quality control.

Deep learning methods for magnetic resonance imaging (MRI) contrast synthesis have advanced substantially, and both conditional generative adversarial networks and regression-based models have shown strong performance in multi-contrast image translation tasks[27-30]. However, most 2D slice-by-slice approaches, often introduce intensity inconsistencies and structural discontinuities across adjacent slices[31,32]. When these predictions are assembled into a 3D volume, the volumetric flickering artifacts may manifest and could compromise the morphological coherence required for reliable segmentation[33,34]. In contrast, 3D volumetric architectures better preserve cross-slice consistency, but they become difficult to train at submillimeter resolution due to computer hardware memory constraints. Recently, diffusion-based synthesis methods can produce high-quality images, but their reliance on iterative denoising processes makes them too computationally expensive for volumetric synthesis in large study cohorts[28,35,36].

In this work, we propose to recover motion-corrupted T2w-TSE images by synthesizing from the concurrently acquired T1w contrast without motion artifact. We present a conditional flow-matching generative model for joint super-resolution and contrast translation from T1w magnetization-prepared rapid gradient-echo (MPRAGE) to T2w-TSE. Specifically, the model was trained using high resolution 7T MRI, tailored for hippocampal subfield analysis. Flow matching enables high-quality synthesis through simple ordinary differential equation (ODE)-based inference by learning velocity fields that map straight paths between noise and target distributions[37], thereby avoiding the slow iterative denoising processes of diffusion models[38]. To preserve volumetric consistency without incurring the memory burden of full 3D architectures, we introduce an autoregressive conditioning scheme in which each synthesized slice conditions the generation of the next. Our proposed model is designed to enable accurate T2w-TSE synthesis from widely available and motion-robust T1w contrast. This directly reduces data loss from motion sensitive T2w-TSE acquisitions and enables hippocampal subfield analysis in datasets where high resolution TSE images are unavailable.

## Results

As illustrated in figure 1, we evaluated the proposed autoregressive conditional flow-matching model across three levels of validation: image synthesis fidelity, hippocampal subfield volumetric agreement in motion free datasets, and clinical sensitivity to Alzheimer's disease-associated atrophy in a large study cohort. First, we assessed whether the synthesized images achieved sufficient visual and quantitative fidelity to support downstream segmentation, including slice-to-slice continuity for anatomy stability. Second, we evaluated whether ASHS-derived subfield volumes from synthesized images agreed with those from as-acquired T2w-TSE scans without motion artifacts. Finally, we tested clinical validity in the ADNI3 cohort, where the as-acquired T2w-TSE images were often severely affected by patient movement. We investigated whether our generative model could recover analyzable data and whether the resulting subfield measurements retained sensitivity to neurodegenerative disease group differences. Our results demonstrate that AI-synthesized images are a viable substitute for directly acquired motion corrupted T2w TSE images, especially in motion-prone cohorts.

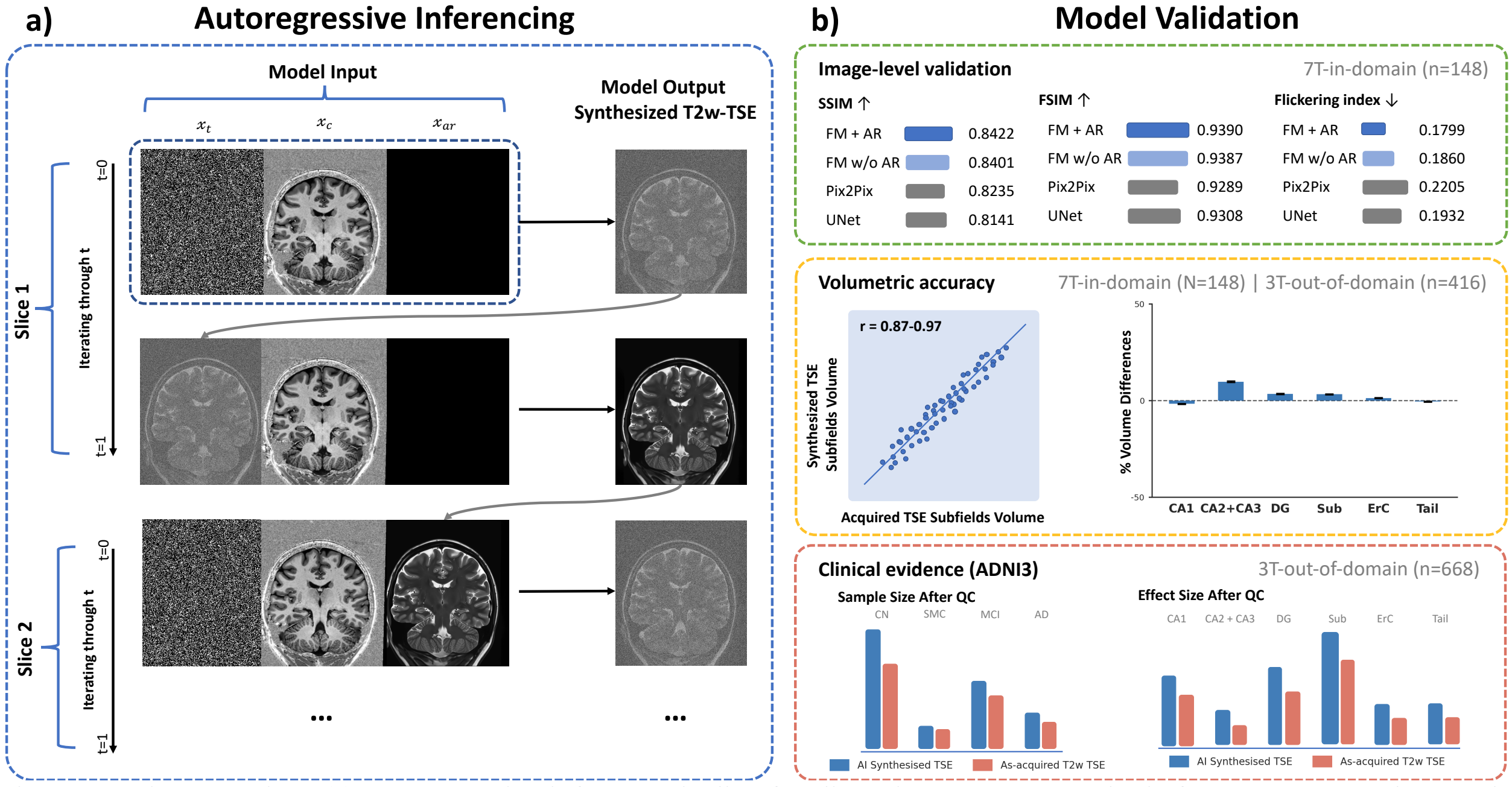

Figure 1 Project overview. (a) Autoregressive inference pipeline for slice-wise T2w TSE synthesis from T1w MRI, where each synthesized slice is fed back as a conditional input for the subsequent slice. (b) Multi-domain validation of synthesis fidelity across three independent cohorts. In the 7T in-domain cohort (n=148), synthesis quality is assessed using image-level metrics including SSIM, FSIM, and Flickering index. Additionally, volumetric accuracy is further validated through hippocampal subfield volume correlation, and subject level volume difference analysis. In a pooled motion artifact-free 3T out-of-domain cohort (n=416) aggregated from multiple open-source datasets, translated contrast fidelity is validated through hippocampal subfield volume correlation and subject level volume difference analysis. In the 3T ADNI3 clinical cohort (n=668), subfield volumes derived from synthesized images demonstrate larger effect sizes in detecting hippocampal atrophy across diagnostic groups (CN, SMC, MCI, AD) compared to as-acquired T2w TSE.

## Visual fidelity and efficient inference

AI-synthesized T2w-TSE images generated from T1w MPRAGE preserved hippocampal anatomy and the internal contrast patterns required for subfield delineation (Fig. 2). Visual inspection showed that the synthesized images retained the characteristic laminar appearance of the hippocampus without obvious structural hallucinations or anatomically implausible features. Notably, this performance was achieved despite the lower native in-plane resolution of the input T1w MPRAGE (1 × 1 mm) relative to the target T2w-TSE acquisition (0.375 × 0.375 mm).

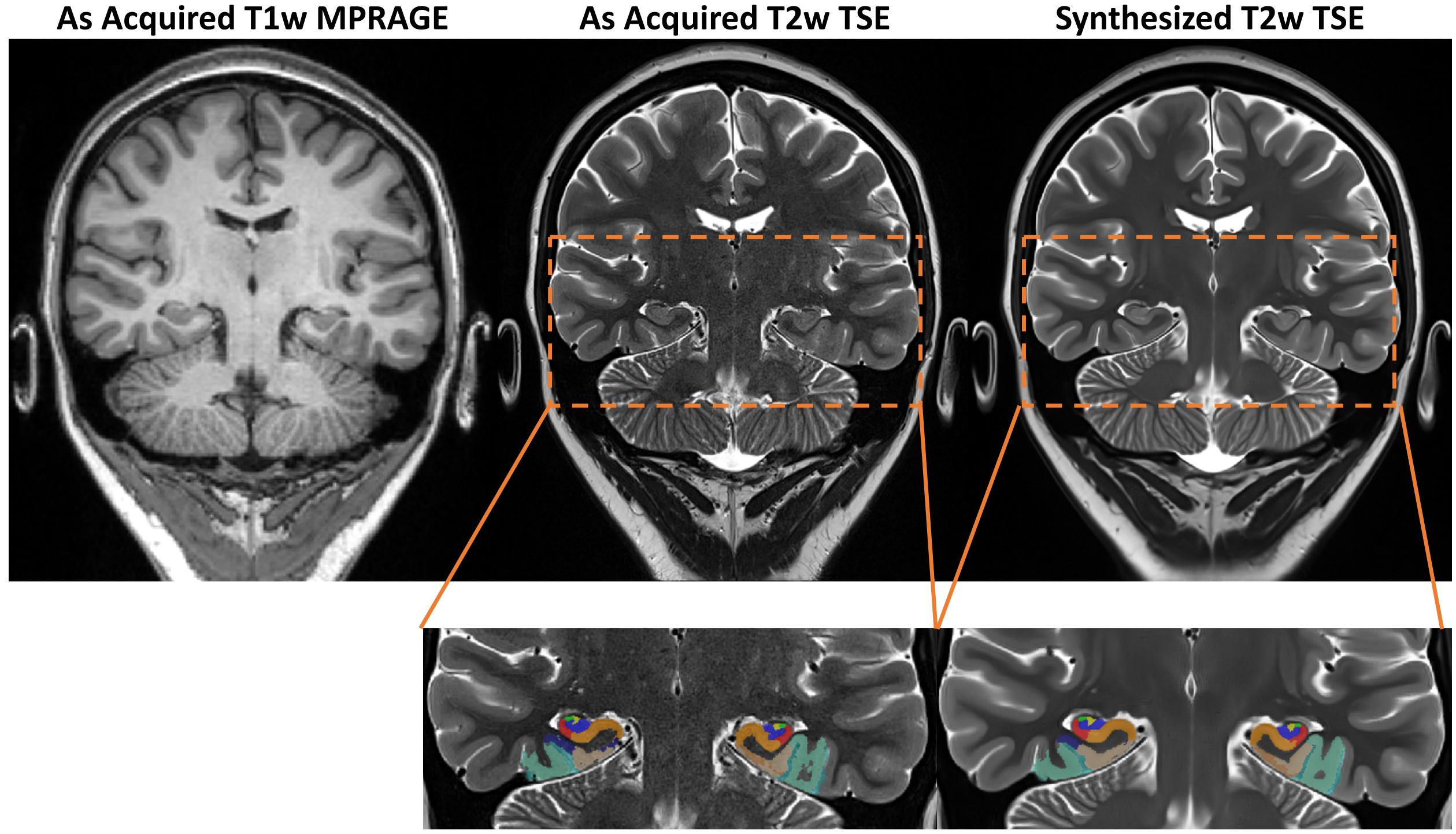

Figure 2 Qualitative example of AI-synthesized T2w-TSE contrast generated from T1w MPRAGE, with corresponding ASHS hippocampal subfield segmentations. The synthesized image preserves hippocampal morphology and internal contrast patterns required for subfield delineation. Segmentations derived from the synthesized images closely match those obtained from high-quality as-acquired T2w-TSE scans.

ASHS-derived hippocampal subfield segmentations from synthesized T2w-TSE images closely resembled those obtained from high-quality as-acquired T2w-TSE scans (Fig. 2). Inference was also computationally efficient: using single-step sampling, the conditional flow-matching model synthesized a full 3D volume of 40 slices in under 30 seconds on an NVIDIA A100 GPU (40 GB), supporting practical use in large-scale research workflows.

### Quantitative image similarity and slice-to-slice continuity

Quantitative results for all models are summarized in Table 1. On the in-domain 7T validation dataset (n=148), we quantified voxel-wise similarity between synthesized and as-acquired images using the structural similarity index (SSIM)[39] and feature similarity index (FSIM)[40]. The proposed autoregressive (AR) flow-matching model achieved an SSIM of 0.8422 ± 0.0802 and a FSIM of 0.9390 ± 0.0239, outperforming the UNet baseline in both SSIM (0.8141 ± 0.0740) and FSIM (0.9308 ± 0.0236), and outperforming Pix2Pix in both SSIM (0.8235 ± 0.0720) and FSIM (0.9289 ± 0.0215). The non-autoregressive flow-matching model performed similarly on these slice-level fidelity metrics (SSIM: 0.8401 ± 0.0800; FSIM: 0.9387 ± 0.0238).

Table 1 Quantitative image fidelity and inter-slice continuity metrics for T2w-TSE synthesis models. Values are mean ± s.d. Lower flickering index indicates better slice-to-slice consistency. Higher SSIM and FSIM indicate greater similarity to the as-acquired T2w-TSE target.

| | **Flow Matching (w/ AR)** | **Flow Matching (w/o AR)** | **Pix2Pix** | **UNet** |
|---|---|---|---|---|
| **Flickering Index** | **0.1799 ± 0.0142** | 0.1860 ± 0.0138 | 0.2205 ± 0.0203 | 0.1932 ± 0.0188 |
| **SSIM** | **0.8422 ± 0.0802** | 0.8401 ± 0.0800 | 0.8235 ± 0.0720 | 0.8141 ± 0.0740 |
| **FSIM** | **0.9390 ± 0.0239** | 0.9387 ± 0.0238 | 0.9289 ± 0.0215 | 0.9308 ± 0.0236 |

Because downstream medical image analysis depends on stable volumetric appearance across adjacent slices, we also assessed slice-to-slice consistency using the flickering index (FI; see Method section). The AR flow-matching model achieved the lowest FI (0.1799 ± 0.0142), indicating the strongest inter-slice continuity, whereas Pix2Pix showed the highest FI (0.2205 ± 0.0203), with UNet (0.1932 ± 0.0188) and the non-autoregressive flow-matching model (0.1860 ± 0.0138) falling in between. Qualitatively, discontinuities were most apparent in low-SNR regions, including the inferior brain and extracranial tissues, where autoregressive conditioning improved anatomical continuity across slices (Fig. 3).

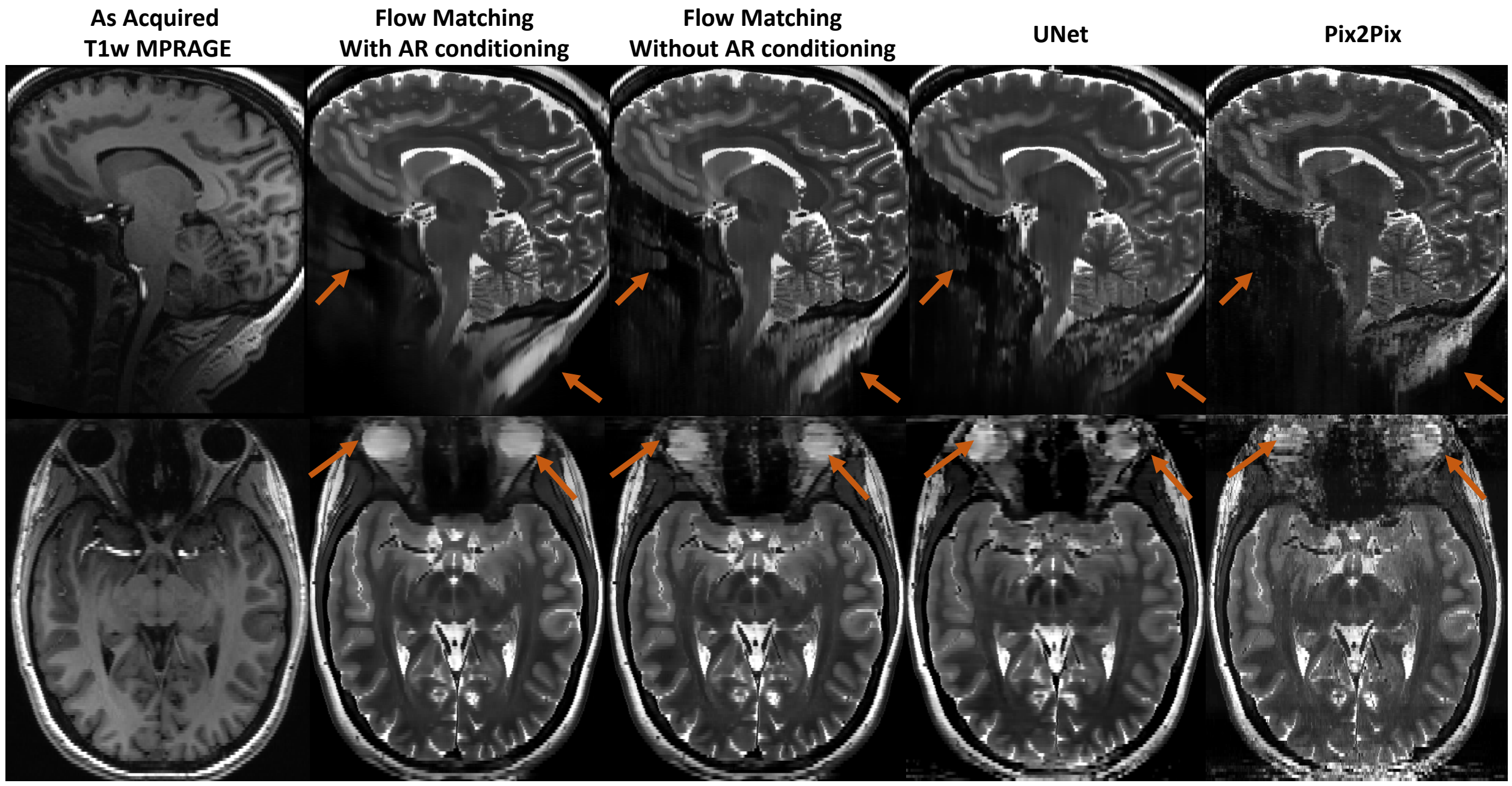

Figure 3 Qualitative example of coronally synthesized T2w TSE volumes across models. Sagittal and axial views show the synthesized results with and without autoregressive conditioning. As pointed out by the arrows, low SNR areas show lower anatomy continuity across slices without autoregressive conditioning. Arrows indicate regions with increased flickering without autoregressive conditioning.

**ASHS subfield volumetric analysis - Internal 7T dataset validation**

Beyond image synthesis fidelity, we further assessed hippocampal subfield segmentation using ASHS on the same 7T validation images. Following segmentation quality control, 117 (left) and 119 (right) subjects were retained. For each subject, volumetric differences were quantified as the percent difference between synthesized and as-acquired T2w-TSE volumes, computed as $(V_{synth} - V_{acq})/V_{acq} \times 100$. Across subfields, subject-level volumetric differences between synthesized and as-acquired T2w-TSE images were generally small, indicating good agreement across subfields (Fig. 4A, Suppl. Table 1). Mean differences were modest across most regions, typically within ±5%. For example, CA1 volumes showed slight decreased volumes (left: −2.76%; right: −3.71%), while DG and Sub exhibited small increased volumes in the synthesized images (DG: left 4.61%, right 4.40%; Sub: left 2.55%, right 1.96%). ErC showed similarly modest increase in volume (left: 2.39%; right: 1.40%), whereas Tail demonstrated small decrease in volume (left: −2.71%; right: −2.58%). Consistent with expectations for smaller structures, CA2+CA3 exhibited larger differences and variability (left: 6.07%; right: 3.52%), likely reflecting increased sensitivity to segmentation variability.

**ASHS subfield volumetric analysis - Public 3T dataset validation**

To assess model generalizability, we applied the model on pooled public 3T datasets containing 416 paired T1w MPRAGE and high-quality T2w-TSE acquisitions without motion artifact. For each subject, we synthesized T2w-TSE contrast from the T1w image and performed ASHS segmentation on both the as-acquired and synthesized images. Following segmentation quality control, 384 (left) and 367 (right) subjects were retained.

Across subfields, ASHS-derived volumes from synthesized images showed strong agreement with those derived from as-acquired T2w-TSE scans (Fig. 4B, Suppl. Table 2), with consistently high Pearson correlations across regions in both hemispheres ($r = 0.873–0.972$). Mean percent differences were generally small, with most regions within ~5%. For example, CA1 differences were minimal (left: −2.84%; right: −0.34%), while DG and Sub showed small increased volumes (DG: left 2.87%, right 4.08%; Sub: left 3.52%, right 3.00%), and ErC and Tail exhibited near-zero differences. The largest deviation was observed in CA2+CA3 (left: 12.45%, right: 6.01%), a relatively small subfield more susceptible to variability in volumetric estimation.

Overall, these results demonstrate that AI-synthesized T2w-TSE images yield hippocampal subfield volumes highly consistent with those derived from as-acquired scans, with only minor regional differences. Most importantly this consistency generalizes from in-domain high-resolution 7T data to more out-of-domain heterogeneous 3T acquisitions.

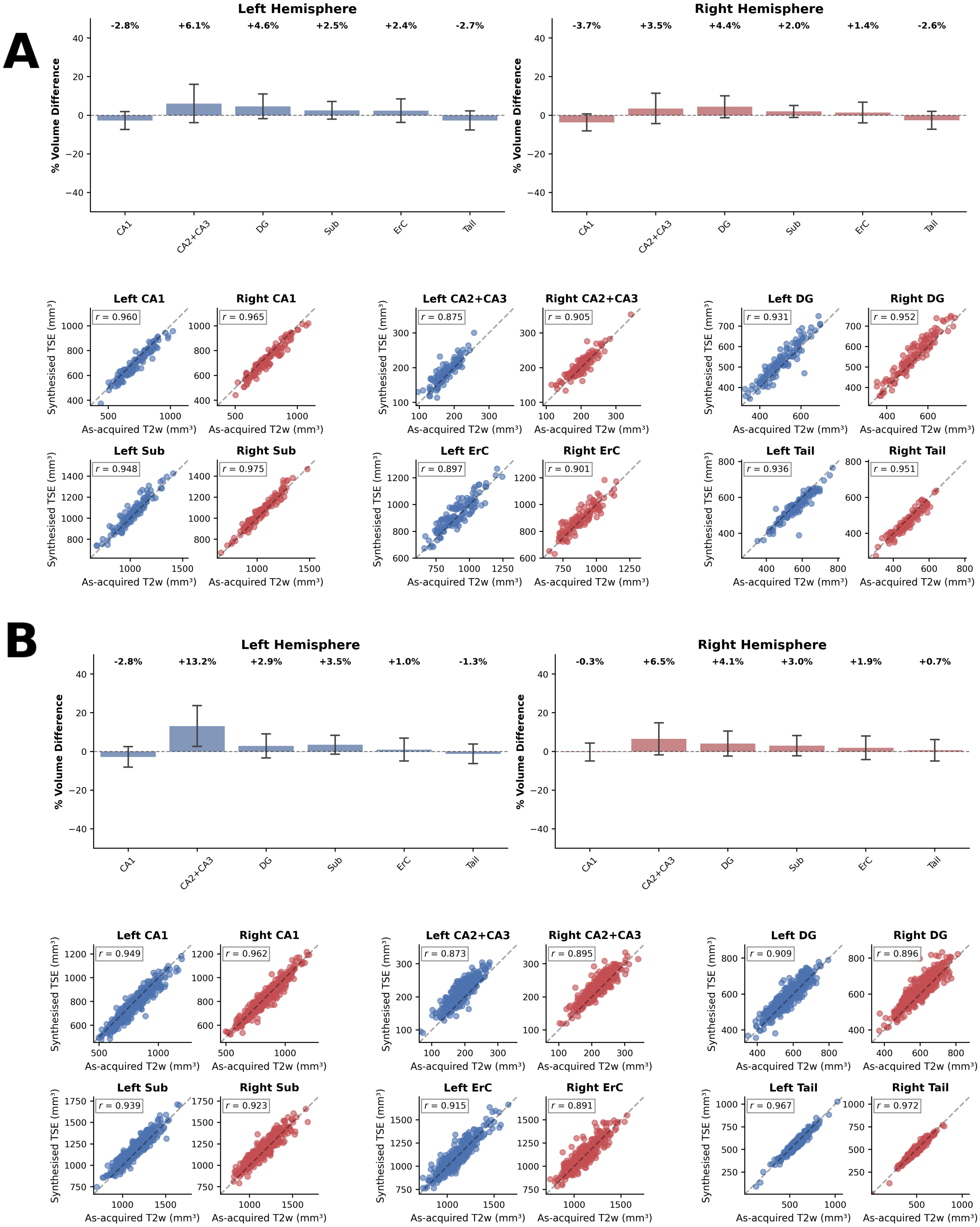

Figure 4 Hippocampal subfield volume agreement between as-acquired and synthesized T2w-TSE images across datasets. (A) Internal 7T dataset; (B) public 3T dataset. For each dataset, the top panel shows bar plots of subject-level volume percent differences derived from ASHS segmentation between as-acquired and synthesized T2w-TSE images across six hippocampal subfields (CA1, CA2+CA3, DG, Sub, ErC, Tail) for left and right hemispheres. Annotations indicate mean percent volume differences. The bottom panel shows scatter plots of per-subject subfield volumes comparing the two image types, with the identity line (dashed) shown for reference. Pearson correlation coefficients are reported for each subfield and hemisphere. Blue and red points represent left and right hemisphere measurements, respectively.

**Clinical validity: Alzheimer's Disease associated subfield atrophy**

Motion corruption in as-acquired T2w-TSE images is a common source of data loss in elderly cohorts. Figure 5 illustrates this directly: motion-corrupted as-acquired T2w-TSE scans exhibit severe ringing and blurring artifacts that obscure hippocampal subfield boundaries, whereas the corresponding AI-synthesized images, generated from motion-free T1w MPRAGE scans, recover clear contrast and well-defined anatomy.

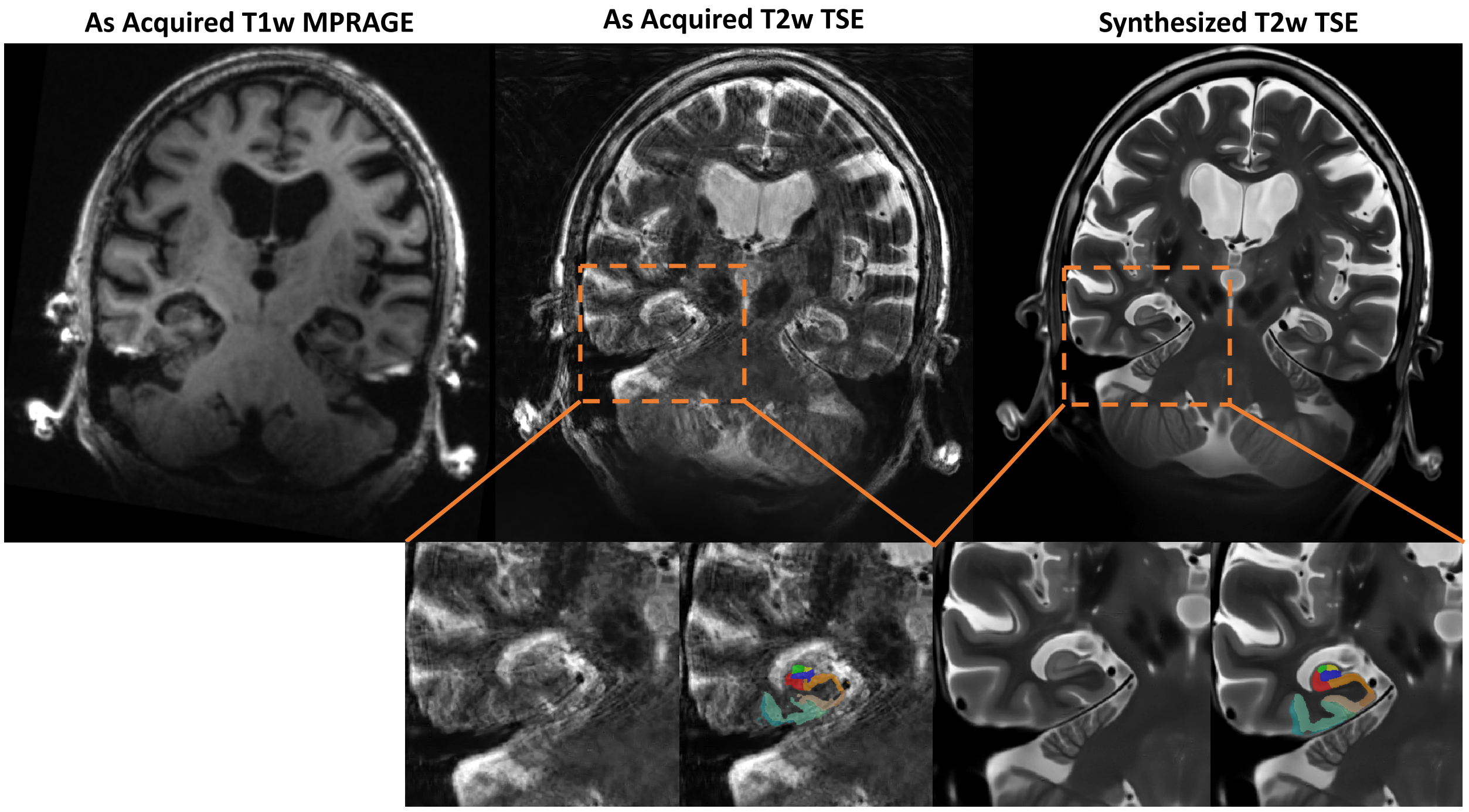

Figure 5 Example of data sample recovery using AI-synthesized T2w-TSE contrast. The as-acquired T1w MPRAGE image is free of visible motion artifact, whereas the corresponding as-acquired T2w-TSE image is motion-degraded and unsuitable for reliable hippocampal subfield segmentation. The AI-synthesized T2w-TSE image generated from the T1w scan shows clear hippocampal subfield delineation and supports anatomically plausible ASHS segmentation, allowing this otherwise excluded case to be retained for analysis.

We evaluated the clinical validity of our model on ADNI3 dataset[7], comprising 668 baseline participants in 4 different groups: cognitively normal (CN; n = 392), subjective memory complaint (SMC; n = 17), mild cognitive impairment (MCI; n = 209), and Alzheimer's disease (AD; n = 50), with a mean age of 72.6 ± 8.2 years across groups.

After quality control, the AI-synthesized TSE image group retained 593 subjects (CN = 350, SMC = 15, MCI = 185, AD = 43), 31.8% larger analyzable cohort than the as-acquired T2w-TSE image group (CN = 281, SMC = 13, MCI = 131, AD = 25; total n = 450; Fig. 6.). Intracranial volume (ICV), age, and sex adjusted subfield volumes showed the expected pattern of progressive atrophy from CN through MCI to AD across both imaging groups. Kruskal-Wallis tests revealed significant group differences (FDR-corrected) in all 7 of 7 subfields in both hemispheres for both the AI-synthesized and as-acquired TSE images. Across all subfields, the AI-synthesized TSE consistently yielded larger effect sizes ($\varepsilon^2$) than the as-acquired image group (Table 2, Fig. 7). The largest effects were observed in the subiculum, which showed the strongest diagnostic group separation in both the left hemisphere (AI-synthesized: H = 98.39, $\varepsilon^2$ = 0.162, p = $5.52\times10^{-20}$; as-acquired: H = 56.86, $\varepsilon^2$ = 0.121, p = $4.40\times10^{-11}$) and right hemisphere (AI-synthesized: H = 81.58, $\varepsilon^2$ = 0.133, p = $1.12\times10^{-16}$; as-acquired: H = 48.58, $\varepsilon^2$ = 0.102, p = $1.28\times10^{-9}$), consistent with known patterns of early AD-related neurodegeneration. Notably, the AI-synthesized images demonstrated particularly pronounced sensitivity gains in CA2+CA3, where effect sizes were more than threefold larger than those of the as-acquired images (left: $\varepsilon^2$ = 0.053 vs. 0.015; right: $\varepsilon^2$ = 0.044 vs. 0.011).

Our results demonstrate that AI-synthesized T2w-TSE images derived from routinely acquired T1w images not only preserve clinically meaningful subfield atrophy patterns in neurodegenerative disease cohort but achieve greater statistical sensitivity than directly acquired high-resolution T2w TSE images.

Table 2 Kruskal-Wallis test results for ICV-, age-, and sex-adjusted hippocampal subfield volumes across diagnostic groups (CN, SMC, MCI, AD) in the ADNI3 cohort. Results are shown for AI-synthesized (AI) and as-acquired (AS) T2w-TSE images. H: Kruskal-Wallis statistic; $\varepsilon^2$: epsilon-squared effect size; p (FDR): Benjamini-Hochberg corrected p-value. Significance: *** p < 0.001, ** p < 0.01, * p < 0.05.

| Hemisphere | Subfield | Mod | H | p (FDR) | ε² | Sig |
|---|---|---|---|---|---|---|
| **Left** | **CA1** | **AI** | **54.43** | **2.43×10⁻11** | **0.087** | *** |
| | | AS | 28.1 | 1.11×10⁻5 | 0.056 | *** |
| | **CA2 + CA3** | **AI** | **33.94** | **2.38×10⁻7** | **0.053** | *** |
| | | AS | 9.79 | 0.022 | 0.015 | * |
| | **DG** | **AI** | **64.15** | **3.05×10⁻13** | **0.104** | *** |
| | | AS | 21.09 | 1.79×10⁻4 | 0.041 | *** |
| | **SUB** | **AI** | **98.39** | **5.52×10⁻20** | **0.162** | *** |
| | | AS | 56.86 | 4.40×10⁻11 | 0.121 | *** |
| | **ERC** | **AI** | **40.25** | **1.51×10⁻8** | **0.063** | *** |
| | | AS | 19.34 | 3.10×10⁻4 | 0.037 | *** |
| | **Tail** | **AI** | **39.45** | **1.86×10⁻8** | **0.062** | *** |
| | | AS | 23.05 | 1.05×10⁻4 | 0.045 | *** |
| | **Whole Hipp.** | **AI** | **74.24** | **2.82×10⁻15** | **0.121** | *** |
| | | AS | 41.29 | 3.03×10⁻8 | 0.086 | *** |
| **Right** | **CA1** | **AI** | **36.76** | **6.37×10⁻8** | **0.057** | *** |
| | | AS | 19.98 | 2.49×10⁻4 | 0.038 | *** |
| | **CA2 + CA3** | **AI** | **28.73** | **2.74×10⁻6** | **0.044** | *** |
| | | AS | 8.11 | 0.0439000 | 0.011 | * |
| | **DG** | **AI** | **49.24** | **2.32×10⁻10** | **0.079** | *** |
| | | AS | 16.64 | 0.0010000 | 0.031 | ** |
| | **SUB** | **AI** | **81.58** | **1.12×10⁻16** | **0.133** | *** |
| | | AS | 48.58 | 1.28×10⁻9 | 0.102 | *** |
| | **ERC** | **AI** | **25.99** | **1.10×10⁻5** | **0.039** | *** |
| | | AS | 12.71 | 0.0061000 | 0.022 | ** |
| | **Tail** | **AI** | **39.52** | **1.86×10⁻8** | **0.062** | *** |
| | | AS | 22.33 | 1.27×10⁻4 | 0.044 | *** |
| | **Whole Hipp.** | **AI** | **62.03** | **6.94×10⁻13** | **0.100** | *** |
| | | AS | 30.76 | 3.82×10⁻6 | 0.062 | *** |

Note: AI rows (blue, bold) = AI-synthesized T2w-TSE; AS rows = as-acquired T2w-TSE. Raw p-values omitted; FDR correction applied using the Benjamini-Hochberg procedure.

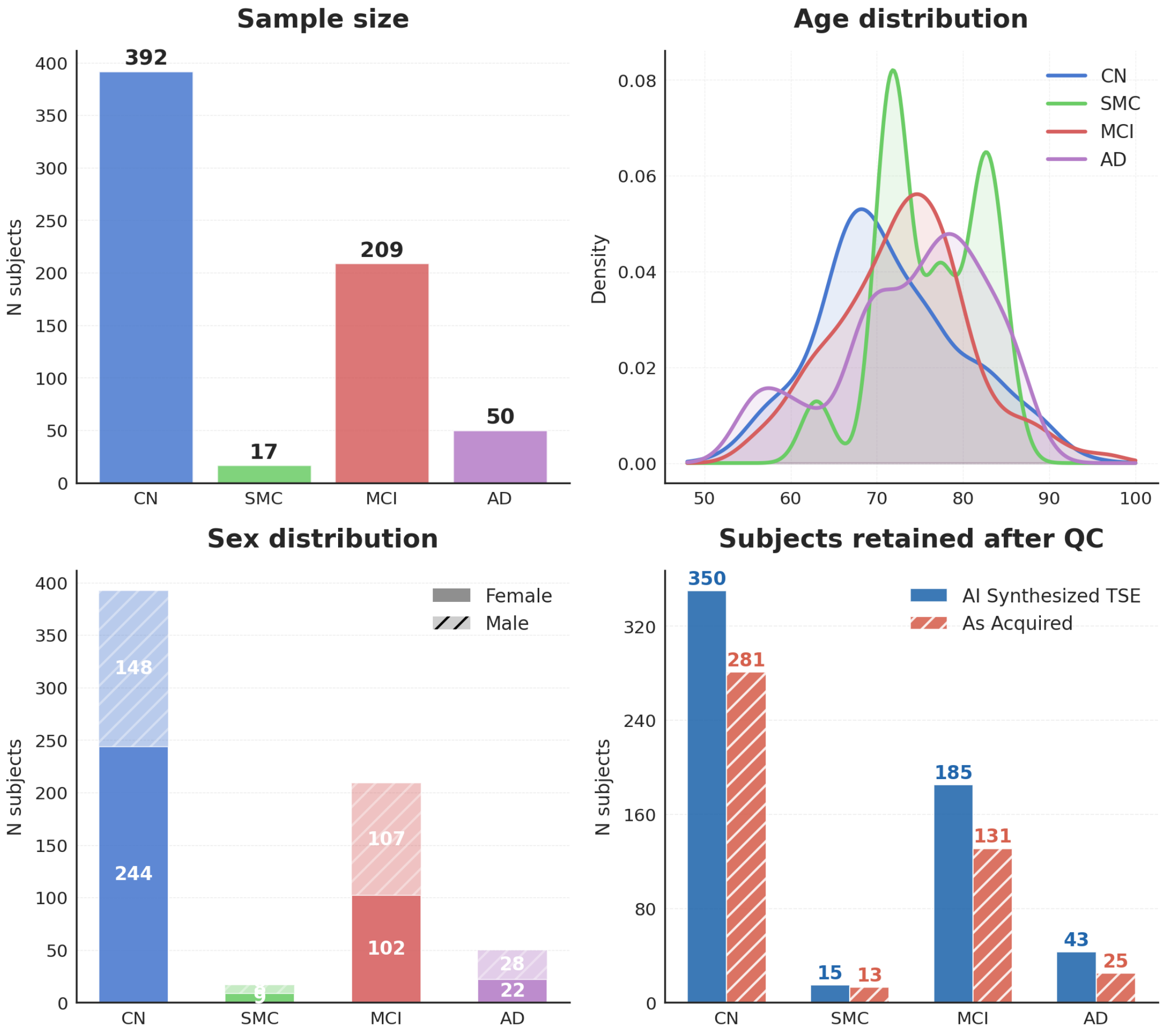

Figure 6 ADNI3 cohort characteristics and ASHS segmentation QC retention. Sample size, age distribution, and sex distribution are shown for 668 baseline participants across four groups: CN (n = 392), SMC (n = 17), MCI (n = 209), and AD (n = 50; mean age 72.6 ± 8.2 years). After segmentation QC, AI-synthesized T2w-TSE images retained 593 subjects compared to 450 for as-acquired T2w-TSE, a 31.8% increase attributable to recovery of motion-corrupted scans. CN, cognitively normal; SMC, subjective memory complaint; MCI, mild cognitive impairment; AD, Alzheimer's disease.

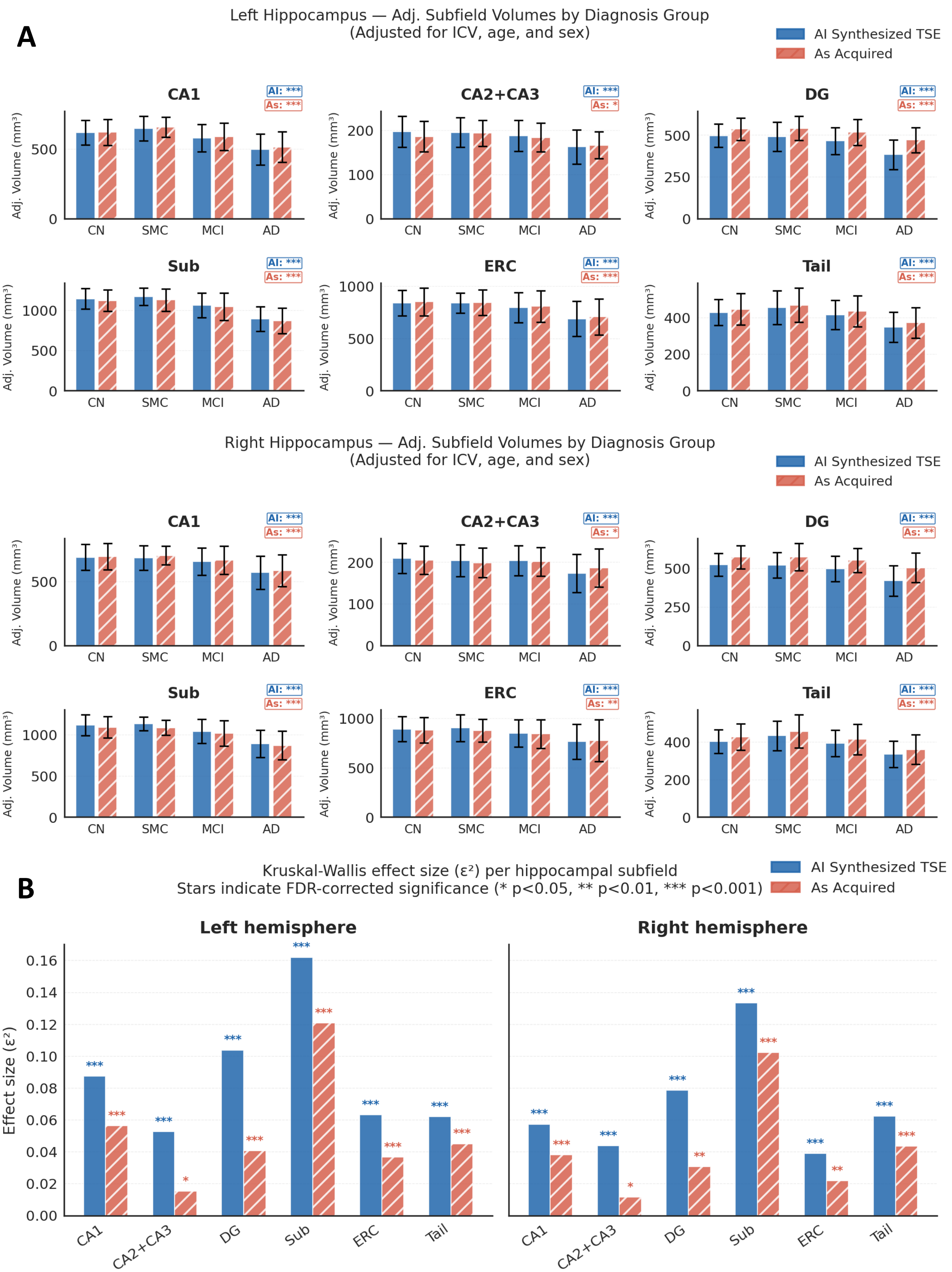

Figure 7 AI-synthesized T2w-TSE images improve sensitivity for detecting hippocampal subfield atrophy across the Alzheimer's disease continuum. (A) ICV, age, and sex-adjusted hippocampal subfield volumes for AI-synthesized TSE (blue) and as-acquired

TSE (red) across four diagnostic groups (CN, SMC, MCI, AD) in the ADNI3 cohort, shown separately for left (top) and right (bottom) hemispheres. Error bars indicate standard deviation. Significance annotations reflect pairwise post-hoc comparisons with FDR correction. (B) Kruskal-Wallis effect sizes ($\varepsilon^2$) per subfield for left and right hemispheres, comparing AI-synthesized and as-acquired T2w-TSE images. Stars indicate FDR-corrected significance of the overall group difference (* $p < 0.05$, ** $p < 0.01$, *** $p < 0.001$). AI-synthesized images consistently yielded larger effect sizes across all subfields and both hemispheres, with the subiculum showing the strongest group separation in both modalities. CN, cognitively normal; SMC, subjective memory complaint; MCI, mild cognitive impairment; AD, Alzheimer's disease.

## Discussion

High-resolution T2w-TSE imaging is the standard sequence for hippocampal subfield delineation, yet its clinical and research utility is fundamentally limited by its sensitivity to motion artifact. In this work, we addressed this problem using a conditional flow-matching model that synthesizes high-quality T2w-TSE contrast directly from routinely acquired T1w images. A central question was whether the synthesized images preserved the quantitative anatomical information required for reliable downstream subfield analysis. Validation on public out of domain 3T datasets spanning multiple scanner manufacturers and a broad range of healthy adults demonstrated strong subfield-level agreement between synthesized and as-acquired images, with Pearson correlations ranging from 0.873 to 0.972 across all regions in both hemispheres and an average whole hippocampus volumetric difference of 1.6%.

With this quantitative foundation established, we next examined whether synthesized images could be substituted for motion-corrupted scans in a clinical research setting while preserving the biological signal of interest. This question is particularly important because motion artifact is among the most common causes of data loss in neuroimaging studies, especially in aging and dementia cohorts, in which motion-sensitive sequences impose a substantial compliance burden on participants with limited tolerance for prolonged acquisitions. Older adults, individuals with greater cognitive impairment, and participants at later disease stages are disproportionately likely to produce motion-degraded MRI scans. As a result, excluding these scans during quality control does not simply reduce statistical power; it may also introduce systematic selection bias by preferentially removing the very participants most likely to exhibit disease-related anatomical abnormalities.

The proposed contrast synthesis model presented in this study directly addresses this problem. By replacing motion-corrupted T2w-TSE images with AI-synthesized counterparts derived from concurrently acquired, motion-resistant T1w images, our approach recovers analyzable hippocampal subfield data from acquisitions that would otherwise be excluded from volumetric analysis. In the ADNI3 cohort, the contrast synthesized image group retained 31.8% more subjects than the as-acquired T2w-TSE group after segmentation quality control, yielding larger effect sizes for Alzheimer's disease-associated subfield atrophy across all seven subfields examined. These findings suggest that AI synthesized contrast can meaningfully increase both the statistical power and the inclusivity of neuroimaging cohort studies, particularly in elderly populations where motion artifact is prevalent.

From a methodological perspective, unlike traditional diffusion models that require iterative denoising, the implemented flow-matching framework allows synthesis through direct integration of a learned velocity field, enabling efficient one-step sampling. When combined with autoregressive conditioning across slices, this framework improves volumetric continuity while maintaining practical computational efficiency. With inference time less than 30 seconds per image volume, it is 1000 times faster than traditional denoising diffusion probabilistic models. Because the method operates on routinely acquired T1w images, it can also be applied retrospectively to existing datasets, potentially enabling hippocampal subfield analysis in cohorts in which dedicated T2w-TSE acquisitions were incomplete, motion corrupted, or never acquired. Across the evaluated baselines, the autoregressive flow-matching model achieved the highest SSIM, FSIM and the lowest flickering index, confirming its advantage in both voxel-level fidelity and volumetric consistency.

Despite these promising results, several limitations should be considered. First, the training data were collected with the acquisition volumes centered on the hippocampal region. As a result, anatomical structures outside this region, including the skull, orbits, and neck musculature, were underrepresented or absent in the training distribution. Consequently, synthesized images show artifacts or reduced fidelity in these peripheral regions (Fig. 3). For this reason, the current model should be used specifically for hippocampal and medial temporal lobe analyses rather than for whole-brain contrast synthesis or global volumetric measurements.

Second, the training cohort was predominantly healthy individuals. The model therefore learns the relationship between T1w and T2w-TSE contrast under a largely normal anatomical prior. Pathological conditions such as tumors, stroke, cortical malformations, or large structural lesions were not represented in the training data. As a result, the model may fail to reproduce unusual structural contrast or may partially suppress features that deviate from the learned distribution. Applications in populations with substantial structural abnormalities should therefore be interpreted with caution until further validation is performed.

Future work should focus on expanding both the anatomical coverage and the population diversity of the training dataset. Increasing the spatial extent of the training acquisitions to include full-brain volumes would allow the model to learn contrast relationships beyond the hippocampal region and may reduce the peripheral artifacts observed in extracranial tissues. Equally important will be the inclusion of more heterogeneous clinical populations. Training data that include individuals with brain tumors, ischemic stroke, and a broader spectrum of neurodegenerative disease would improve the model's ability to generalize to pathological anatomy and reduce the risk of synthesis bias.

## Method

### Model architecture

We repurposed MONAI's 2D DiffusionModelUNet[41] as the backbone for flow matching-based contrast synthesis, adapting it for velocity prediction rather than stochastic noise prediction. The architecture consists of a U-Net encoder-decoder with three feature levels (256, 256, and 512 channels), and self-attention at the third level (figure 8). The model processes 3-channel inputs: the intermediate noisy sample $x_t$, the conditioning slice $x_c$ (e.g. T1w image), and the autoregressive conditioning slice $x_{ar}$. This design stabilizes and simplifies pixel-level flow matching model training while ensuring global volumetric consistency and preserving computational efficiency relative to 3D-based models.

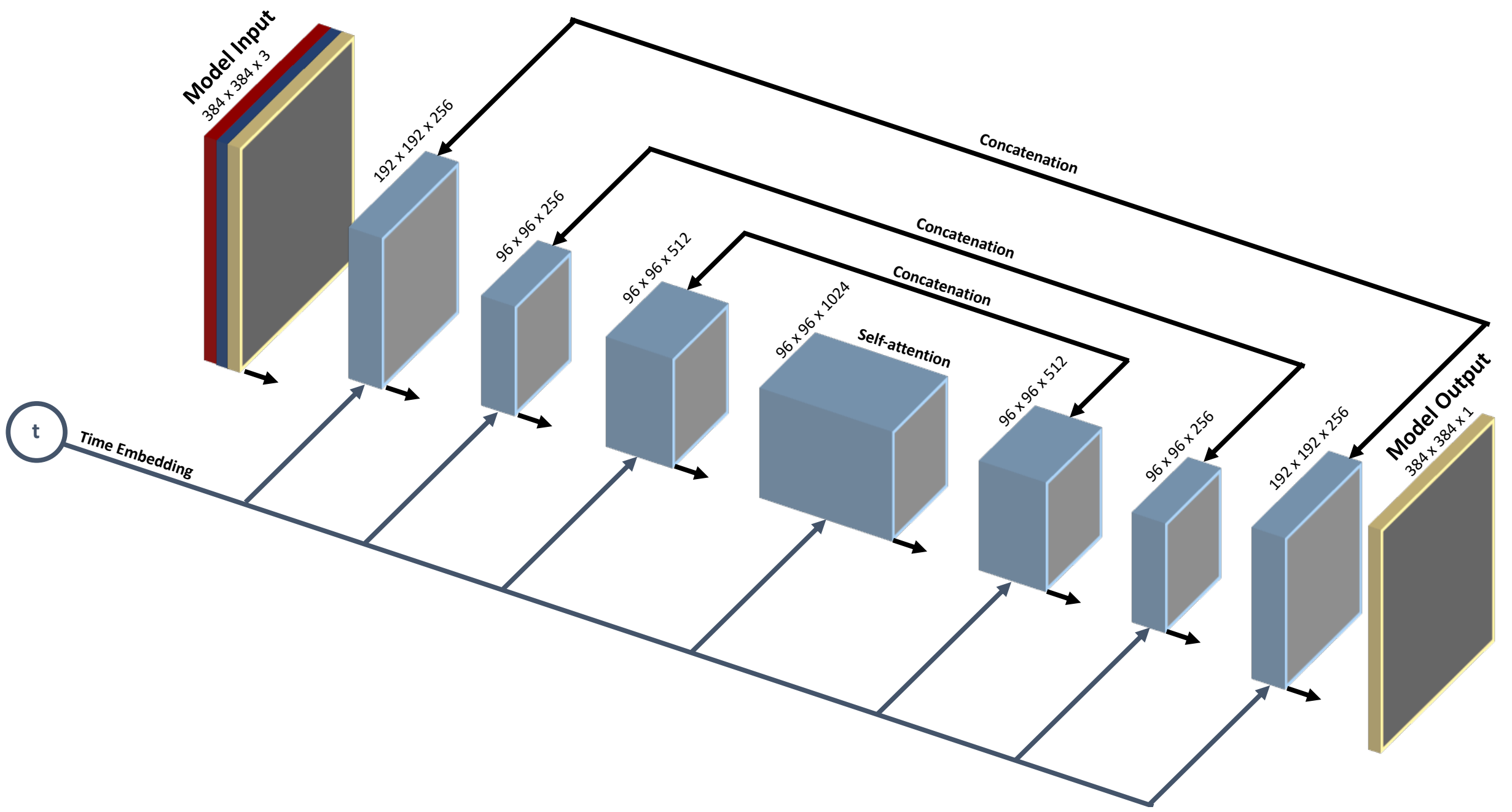

Figure 8 Model architecture overview. The flow-matching model uses a 2D U-Net backbone with three feature levels (256, 256, and 512 channels) and self-attention at the deepest level.

**Baseline comparison**

To contextualize the performance of the proposed model, we trained two additional baseline models using the same DiffusionModelUNet architecture as the proposed autoregressive flow-matching model, ensuring that architectural differences alone do not explain performance variations. For both baselines, the time embedding was fixed to zero during training and inference, effectively disabling any time conditioning and reducing the model to a standard image-conditional UNet.

Baseline: the UNet model was trained with a standard pixel-wise mean squared error (MSE) loss directly between the input T1w image and the target T2w-TSE image, without any adversarial or flow matching training objective. During training, UNet model was optimized using the Adam optimizer with a learning rate of 3e-5.

Pix2Pix: the same UNet model was paired with a 70x70 PatchGAN discriminator to form a Pix2Pix model[42], trained with the standard adversarial and MSE loss combination. The discriminator takes the concatenation of the conditioning T1w slice and the real or synthesized T2w-TSE slice as input and produces patch-level real/fake predictions. To stabilize training, the discriminator was not activated for the first 10 epochs, during which the generator was trained with MSE loss only. The generator and discriminator were optimized separately using the Adam optimizer, where the generator had a learning rate of 3e-5, and the discriminator had a learning rate of 1e-5.

Flow Matching without autoregressive conditioning: to isolate the contribution of autoregressive conditioning, we evaluated the trained flow-matching model with the third input channel $x_{ar}$ fixed to a zero tensor at inference, effectively removing cross-slice context. No separate training was required. This baseline serves as an inference-time ablation to quantify the impact of autoregressive conditioning on slice-to-slice consistency.

All models were trained for 100 epochs on 8 NVIDIA A100 GPUs (40 GB) using Distributed Data Parallel (DDP). During training, images were cropped online from 512×512 to 384×384 to reduce memory overhead and support a total batch size of 32.

**Flow matching training objective with autoregressive conditioning**
We implemented the flow matching training objective following[37]. We adapted the denoising diffusion model from MONAI and incorporated autoregressive conditioning for enhanced cross-slice consistency. Specifically, given a noisy source $x_0$ and a clean target image $x_1$ the linear interpolation with small Gaussian perturbation can be written as:

$$x_t = (1 - t)x_0 + tx_1, \ t \in [0,1],$$

Under this interpolation, the velocity field at time t is:

$$v_t = \frac{d}{dt}x_t = -x_0 + x_1$$

Let $v_\theta$ denote the neural network parameterized by $\theta$, which takes a three channel input: $x_t, x_c, x_{ar}$ , where $x_t$ is the intermediate noisy version of the T2w-TSE image, $x_c$ is the T1w image, and $x_{ar}$ is the T2w-TSE image of the previous frame. We train the model $v_\theta$ to minimize:

$$\mathcal{L} = \mathbb{E}[||v_\theta(x_t, x_c, x_{ar}), (-x_0 + x_1)||_2^2]$$

**Inference via ordinary differential equation sampling**
At inference, synthesis is performed by integrating the learned velocity field using Euler's method. The flow matching framework transforms the image contrast synthesis problem into solving a continuous-time system, where the learned velocity field guides the transformation from noise to the target contrast. The synthesized image $x_1$ can be obtained with integration: $x_1 = x_0 + \int_0^1 v_\theta(x_t, x_c, x_{ar})\, dt$, with $x_0$ being the gaussian noise. All T1w structural image was first co-registered to the T2w-TSE image. The registered T1w image was then fed into the model for contrast synthesis.

**Training dataset**
We curated 725 paired T1w and T2w-TSE volumes from our in-vivo imaging bank at the University of Pittsburgh for supervised contrast translation training. The dataset spans a wide age range (16–96 years) and includes two T1w acquisition protocols: 158 pairs of 0.55 mm isotropic T1w MP2RAGE with T2w-TSE, and 567 pairs of 0.75 mm T1w MPRAGE with T2w-TSE. Of these, 577 pairs were used for training and the other 148 pairs were held out for quantitative evaluation of voxel-level metrics (SSIM, FSIM, and FI). All acquired T2w-TSE images are of resolution 0.375x0.375x1.5 mm. All acquisitions were performed on a 7T Magnetom scanner (Siemens Healthineers) using the in-house developed TAC RF-coil family[43-45], with participants' informed consent and approval from the University of Pittsburgh Institutional Review Board. The curated images were free of motion artifacts.

All T1w images were rigidly registered to the T2w-TSE space and cropped or zero-padded to have an in-plane size of 512x512. Bias field correction was applied to all images using SPM12[46].

Volumes were then sliced along the in-plane axis. To improve robustness for real-world acquisition variability, online data augmentation was performed using TorchIO[47], including simulated ghosting, motion, bias field, and noise artifacts.

**Public 3T validation dataset**

To assess generalizability across field strengths, scanner manufacturers, and populations, we evaluated the model on publicly available datasets from OpenNeuro (IDs: ds001517, ds001946, ds002168, ds002813, ds003707, ds003778, ds003851, ds004094, ds004349, ds005947, ds006039)[48-58], comprising 417 subjects with paired T1w MPRAGE (0.8–1 $mm^3$ isotropic) and coronal T2w-TSE images (in-plane resolution 0.44–0.5 mm, slice thickness 1.5–4 mm, acquired perpendicular to the hippocampal axis). These datasets were acquired on 3T scanners (Siemens Prisma, Skyra, and GE Healthcare) using 20–64 channel head coils, covering primarily healthy young adults and a subset with psychiatric conditions. Images were of high quality with minimal motion artifacts, providing a clean benchmark for downstream ASHS hippocampal subfield segmentation accuracy assessment. Subfield segmentations on the as acquired T2w TSE images and on the AI synthesized images were performed using ASHS with the Magdeburg atlas[59]. Hemispheric segmentation quality control was applied to all 417 subjects; after exclusions, 384 and 367 hemispheres were retained for left and right subfield analysis, respectively. To assess the volumetric agreement between the as acquired and AI-synthesized images, paired Student's t-tests were performed for each subfield in each hemisphere, with p-values corrected for multiple comparison. Pearson correlation coefficients were additionally computed per subfield to quantify the linear correspondence between subfield volumes derived from the images.

**Alzheimer's disease clinical dataset**

We further evaluated our model using the Alzheimer's Disease Neuroimaging Initiative 3, a multicenter study designed to develop imaging and molecular biomarkers for the early detection and tracking of Alzheimer's disease[7]. The study cohort comprised baseline visits from 668 participants across four groups: cognitively normal (CN; n=392, 148 male/244 female), subjective memory complaint (SMC; n=17, 8 male/9 female), mild cognitive impairment (MCI; n=209, 107 male/102 female), and Alzheimer's disease (AD; n=50, 28 male/22 female). Only the baseline visits imaging data was used in this study to ensure independence between observations.

ADNI3 imaging was exclusively acquired at 3T across GE, Philips, and Siemens scanners at 57 imaging centers. T1w MPRAGE and the T2w TSE images were used for this study. Specifically, the T1w structural image is of 1 $mm^3$ isotropic resolution and the T2w TSE is of 0.39x2x0.39 mm resolution acquired in the oblique coronal orientation, perpendicular to the hippocampal axis. Subfield segmentation was performed using ASHS with the 3T Penn Memory Center atlas[60] for both the as acquired T2w TSE images and the AI synthesized images. Following segmentations, all outputs underwent subject level quality control. Subfield volumes were adjusted for intracranial volume, age, and sex prior to statistical analysis. Subfield volume group differences were assessed using the Kruskal-Wallis test, with p-values corrected for multiple comparison. Effect size was quantified using epsilon-squared ($\varepsilon^2$).

**Pixel-level image quality evaluation**

Synthesized image quality was assessed using structural similarity index and feature similarity index against the as-acquired T2w-TSE reference. Because rigid registration cannot guarantee perfect voxel-wise alignment in extracerebral regions, all images were skull-stripped with mri_synthstrip[61] prior to computing SSIM and FSIM to restrict evaluation on anatomically meaningful regions where registration is reliable.

To quantify volumetric coherence across the synthesized 2D stack, we computed a flickering index (FI) across adjacent slices. This metric captures undesirable intensity discontinuities that can arise when slices are synthesized independently without cross-slice context. FI is defined as: $FI = \frac{\mathbb{E}[|I_{n+1} - I_n|]}{\mu}$, where $\mu$ is the mean pixel intensity of the 2D image stack and $\mathbb{E}[|I_{n+1} - I_n|]$ is the average absolute intensity difference between adjacent slices across the volume. Lower FI values indicate greater slice-to-slice consistency.

### Quality control

Image quality control was performed in two stages prior to downstream analysis. In the first stage, all acquired T2w-TSE images underwent visual inspection using a strict binary criterion: images were classified as either pass or fail based on the presence of motion artifacts. Images were assigned a passing grade only when the hippocampal region was entirely free of motion-related signal degradation. Images exhibiting artifacts of any severity, including minor ghosting or blurring, were excluded to ensure the validity of comparisons. In the second stage, segmentation outputs produced by ASHS were reviewed independently. Segmentations were flagged for exclusion when critical subregions were incorrectly delineated, such as partial omission of the CA1 subfield boundary. Representative examples of excluded T2w-TSE images and erroneous ASHS segmentations are provided in Suppl. Fig. 1 and Suppl. Fig. 2, respectively.

### Ethics statement

All procedures performed in this study involving human participants were in accordance with the Declaration of Helsinki. The collection and use of the 7T MRI data were approved by the University of Pittsburgh Institutional Review Board (approval number: STUDY19030338). Written informed consent was obtained from all participants. The publicly available 3T datasets used in this study (OpenNeuro, ADNI3) were collected under protocols approved by their respective institutional review boards, with written informed consent obtained from all participants.

## Data Availability

The 7T MRI data used for training in this study are not publicly available due to institutional restrictions and research participant privacy protections. However, the 3T data used for evaluation are publicly accessible: the validation dataset is available at https://openneuro.org/, and the ADNI3 dataset can be accessed via the Alzheimer's Disease Neuroimaging Initiative at https://adni.loni.usc.edu/data-samples/adni-data/. The code will be organized at: https://github.com/jinghangli98/MRecover.

## Author Contributions

J.L. contributed to conceptualization, methodology, data acquisition, data analysis and interpretation, software, validation, writing – original draft, writing – review and editing, and visualization. T.S. contributed to methodology, data analysis and interpretation, software, writing

– review and editing, and supervision. C.C. (1) contributed to data analysis and interpretation and writing – review and editing. B.A. contributed to data acquisition and data analysis and interpretation. C.C. (2), S.A., A.S., J.B.(1), H.J., T.C., contributed to data acquisition and data analysis and interpretation. J.B. (2) contributed to resources, writing – review and editing, and project administration. J.M. contributed to investigation and writing – review and editing. A.G., H.A., M.W., contributed to investigation, resources, and writing – review and editing. T.S.I. contributed to conceptualization, methodology, data analysis and interpretation, investigation, writing – review and editing, supervision, project administration, and funding acquisition.

## Competing Interest

All authors declare no competing financial or non-financial interests.

## Acknowledgement

This study was supported by grants R01MH111265, R01AG063525, R01AG055389, R01AG085566, R56AG074467. This research was supported in part by the University of Pittsburgh Center for Research Computing and Data, RRID:SCR_022735, through the resources provided. Specifically, this work used the HTC cluster, which is supported by NIH award number S10OD028483. This research was supported in part by the University of Pittsburgh Center for Research Computing and Data, RRID:SCR_022735, through the resources provided. Specifically, this work used the H2P cluster, which is supported by NSF award number OAC-2117681. We sincerely thank our University of Pittsburgh's 7 Tesla Bioengineering Research Program (7TBRP) collaborators (https://www.7tbrp.pitt.edu/about/collaborators) for their invaluable contributions. We also acknowledge the MRI technicians, students, and research staff at the 7TBRP for their efforts in acquiring the 7T training dataset.

# Supplementary Materials

Supplementary Material for: “*MRecover*: A Conditional Generative Model for Recovering Motion-Corrupted MR images Using AI Generated Contrast”

Supplementary Table 1. Comparison of hippocampal subfield volumes derived from as-acquired and AI-synthesized T2w-TSE images in the internal 7T validation dataset. For each region, the table reports the sample size (after segmentation quality check), mean volume for as-acquired and synthesized images, subject level percent difference (Δ%), and Pearson correlation coefficient (r).

| Hemisphere | Region | N | As acquired TSE mean | AI Synthesized TSE mean | Δ% | r |
|---|---|---|---|---|---|---|
| left | CA1 | 117 | 713.12 | 693.09 | -2.76 | 0.960 |
| left | CA2+CA3 | 117 | 174.88 | 184.97 | 6.07 | 0.875 |
| left | DG | 117 | 501.58 | 523.67 | 4.61 | 0.931 |
| left | Sub | 117 | 1025.74 | 1051.04 | 2.55 | 0.948 |
| left | ErC | 117 | 907.71 | 927.40 | 2.39 | 0.897 |
| left | Tail | 117 | 566.46 | 550.06 | -2.71 | 0.936 |
| right | CA1 | 119 | 763.56 | 735.32 | -3.71 | 0.965 |
| right | CA2+CA3 | 119 | 200.04 | 206.08 | 3.52 | 0.905 |
| right | DG | 119 | 533.27 | 556.37 | 4.40 | 0.952 |
| right | Sub | 119 | 1043.38 | 1063.45 | 1.96 | 0.975 |
| right | ErC | 119 | 883.24 | 893.97 | 1.40 | 0.901 |
| right | Tail | 119 | 479.55 | 466.55 | -2.58 | 0.951 |

Supplementary Table 2. Comparison of hippocampal subfield volumes derived from as-acquired and AI-synthesized T2w-TSE images in the public 3T dataset. For each region, the table reports the sample size (after quality check), mean volume for as-acquired and synthesized images, subject level percent difference (Δ%), and Pearson correlation coefficient (r).

| Hemisphere | Region | N | As acquired TSE mean | AI Synthesized TSE mean | Δ% | r |
|---|---|---|---|---|---|---|
| left | CA1 | 384 | 797.02 | 773.99 | -2.84% | 0.949 |
| left | CA2+CA3 | 384 | 186.44 | 209.65 | 13.16% | 0.873 |
| left | DG | 384 | 566.45 | 581.47 | 2.87% | 0.909 |
| left | Sub | 384 | 1127.21 | 1164.75 | 3.52% | 0.939 |
| left | ErC | 384 | 1099.03 | 1108.13 | 0.97% | 0.915 |
| left | Tail | 384 | 579.92 | 571.95 | -1.25% | 0.967 |
| right | CA1 | 367 | 833.67 | 830.26 | -0.34% | 0.962 |
| right | CA2+CA3 | 367 | 214.58 | 227.48 | 6.52% | 0.895 |
| right | DG | 367 | 592.52 | 615.28 | 4.08% | 0.896 |
| right | Sub | 367 | 1147.67 | 1179.52 | 3.00% | 0.923 |
| right | ErC | 367 | 1089.67 | 1108.17 | 1.89% | 0.891 |
| right | Tail | 367 | 494.67 | 496.60 | 0.67% | 0.972 |

Supplementary Figure 1. Representative example of image quality control. Only images meeting high-quality standards were assigned a passing grade, while those with moderate or severe motion artifacts were excluded from hippocampal subfield analysis.

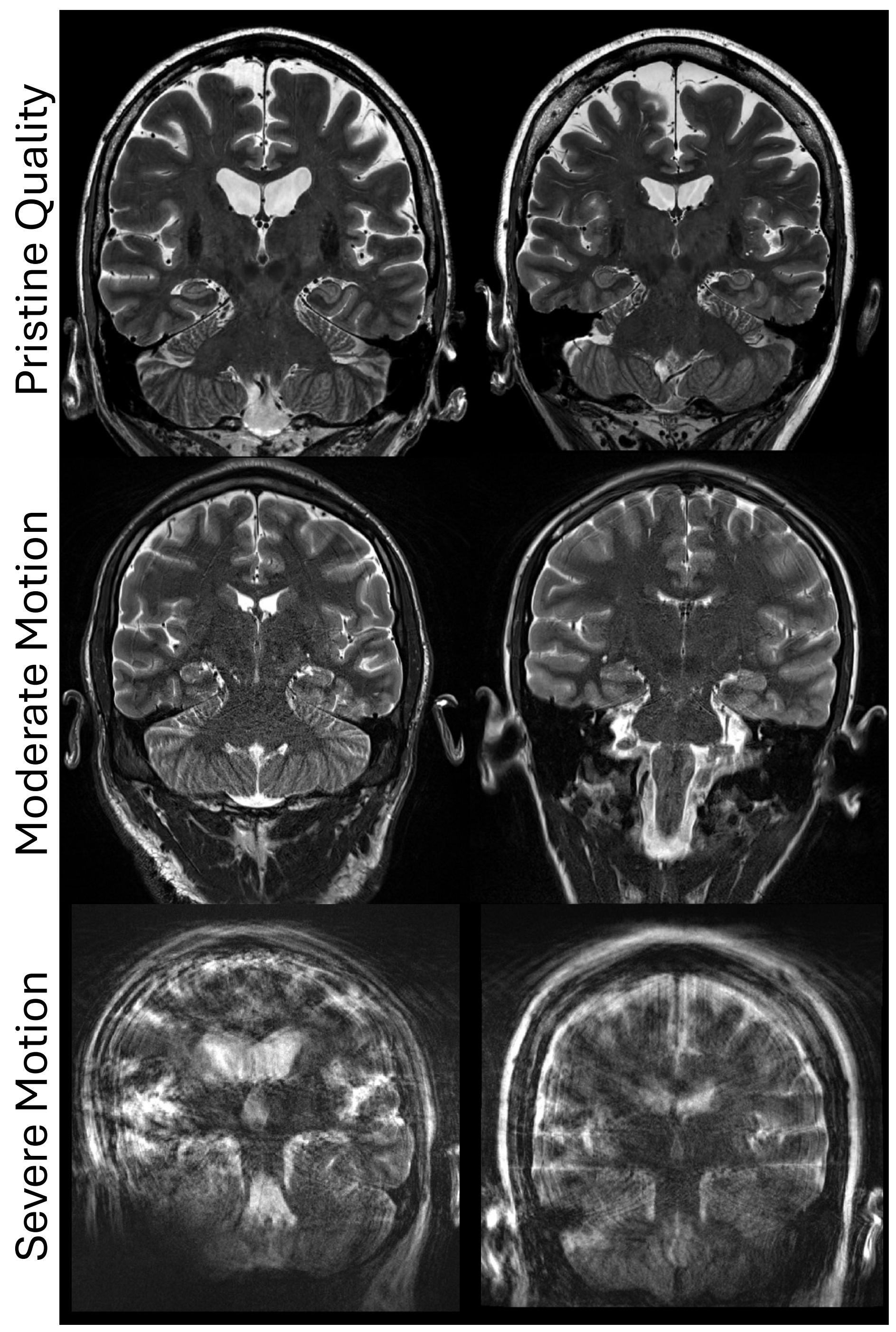

Supplementary Figure 2. Examples of excluded hippocampal subfield segmentations. Only high-quality segmentation outputs were included in the final analysis. The two examples illustrate typical segmentation failures: (i) missed segmentation at the CA1/CA2 boundary and (ii) erroneous leakage of CA1 into the neighboring cerebrospinal fluid.

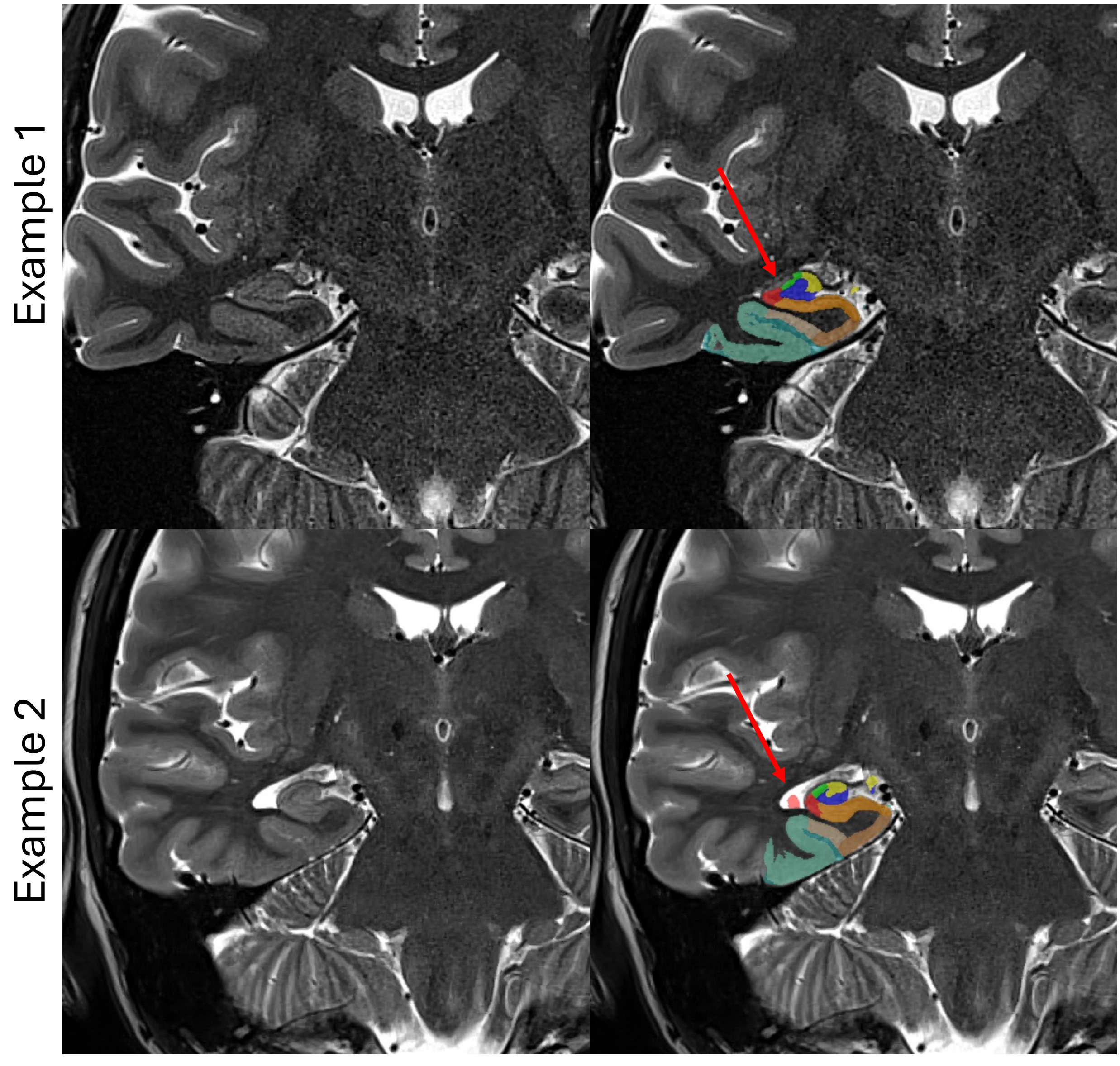
Example 1
Example 2